\documentclass[11pt,a4paper]{article} 
\usepackage{authblk}
\usepackage[hyperref]{acl2021}
\usepackage{times}
\usepackage{latexsym}

\usepackage{multirow}
\usepackage{tabularx}
\usepackage{graphicx}
\usepackage{graphics}

\usepackage{microtype}

\aclfinalcopy 

\title{Towards Knowledge-Grounded Counter Narrative Generation \\ for Hate Speech}

\author[1,2]{Yi-Ling Chung}
\author[1]{Serra Sinem Tekiro\u{g}lu}
\author[1]{Marco Guerini}
\affil[1]{Fondazione Bruno Kessler, Via Sommarive 18, Povo, Trento, Italy }
\affil[2]{University of Trento, Italy \protect\\ 
\protect\\ \texttt{ychung@fbk.eu, tekiroglu@fbk.eu, guerini@fbk.eu}
 }

\begin{document}
\maketitle
\begin{abstract}
Tackling online hatred using informed textual responses -- called counter narratives -- has been brought under the spotlight recently. Accordingly, a research line has emerged to automatically generate counter narratives in order to facilitate the direct intervention in the hate discussion and to prevent hate content from further spreading. Still, current neural approaches tend to produce generic/repetitive responses and lack grounded and up-to-date evidence such as facts, statistics, or examples. Moreover, these models can create plausible but not necessarily true arguments. In this paper we present the first complete knowledge-bound counter narrative generation pipeline, grounded in an external knowledge repository that can provide more informative content to fight online hatred. Together with our approach, we present a series of experiments that show its feasibility to produce suitable and informative counter narratives in in-domain and cross-domain settings.
\end{abstract}

\section{Introduction}
Standard approaches for online hate mitigation generally rely on content moderation, ranging from deletion of hate content and suspension of user accounts to shadow banning. However, these approaches may draw limits on free speech and diverse opinions. An alternative approach is to directly intervene in the conversation with counter narratives. A Counter Narrative (\texttt{CN}) is a non-negative response to a Hate Speech (\texttt{HS}), targeting and contradicting extreme statements with fact-bound arguments or alternative viewpoints \cite{benesch2014countering, schieb2016governing}. Such strategy seeks to de-escalate the conversation, disengage from hateful sentiment and encourage mutual understanding through exchange of opinions. Many Non-Governmental Organizations (NGOs) specialized in hate countering are already adopting this approach by training operators to compose counter narratives. According to NGO guidelines\footnote{For example:  \url{https://getthetrollsout.org/stoppinghate}}, proper \texttt{CN}s should also include credible evidence.

\begin{table}[ht!]
\small
  \centering
  \begin{tabular}{p{0.95\linewidth}}
    \hline
\textbf{HS:} \textit{The world would be a better place without Muslims. They are only killing and raping our children.} \\
\hline
\textbf{CN1:} \textit{Evidence for this? This is not true. How can you say this about an entire faith?} \\
\hline
\textbf{CN2:} \textit{On the contrary, most children abuse is operated by people they know: a relative, family friends, sports coach, someone in a position of trust and authority. Besides, Muslims help people - A Muslim woman rushed to help the victims of a triple stabbing in Manchester on New Year’s Eve.} \\ 
\hline
\textbf{CN3:} \textit{You are truly one stupid backwards thinking idiot to comment on Muslims like that.} \\
    \hline 
  \end{tabular}
  \caption{An example hate speech with appropriate counter narratives (\texttt{CN1} and \texttt{CN2}) and an inappropriate one (\texttt{CN3}) containing abusive language.}
 \label{tab:CN-example}
\end{table}

In Table \ref{tab:CN-example}, we present an \texttt{HS} along with several \texttt{CN}s. Although \texttt{CN1} and \texttt{CN2} are both appropriate responses, not all appropriate \texttt{CN}s are equally effective \cite{silverman2016impact, tuck2016counter}. \texttt{CN2} is expected to be more effective as it is tailored to the \texttt{HS} and demonstrates contextualized and persuasive statements with supporting facts, knowledge, and logical reasoning \cite{habernal2016argument} rather than a simple generic request for evidence as in \texttt{CN1}. Conversely, \texttt{CN3} is considered as a natural but inappropriate and aggressive response.

In this context, scaling manual response to massive amounts of online hatred is an undeniably challenging task for NGO operators.
Recently, studies have started to focus on data and strategies for the automatic generation of counter narratives in an end-to-end fashion \cite{chung2019conan,qian2019benchmark, tekiroglu2020generating, chung2020italian}. 
However, the seminal studies, based on training models using only the \texttt{HS}-\texttt{CN} data samples, do not directly address the required properties for proper and effective \texttt{CN}s such as providing credible evidence and knowledge with facts, statistics or examples. While such models possess knowledge to a certain extent through utilizing existing generative language models (LMs) such as GPT-2 \cite{radford2019language}, this knowledge (e.g., events or news) will get out-dated over time. Furthermore, such models are able to create plausible but not necessarily true arguments - a problem known as content hallucination - \cite{zellers2019defending, solaiman2019release}. A hallucinated example of a fact-bound \texttt{CN}, generated by GPT-2 model fine-tuned as done by \citet{tekiroglu2020generating}, is provided in Table \ref{tab:CN-wrong}. 
Therefore, in this paper we investigate the generation of knowledge-bound counter narratives, which had never been studied yet to the best of our knowledge. We hypothesized that knowledge infusion can not only leverage generating informative \texttt{CN}s but also handling hate speech from unseen domains (i.e. hate targets) without further training (e.g., train on Islamophobia and test on Antisemitism).

\begin{table}[ht!]
\small
  \centering
  \begin{tabular}{p{0.95\linewidth}}
    \hline
\textbf{HS:} \textit{Muslims are invading our country} \\
\hline
\textbf{CN:} \textit{Actually, there were 16,938,000 Muslims in the UK in 2016, so if you exclude London, that is actually increasing by 2\%, which doesn't seem very significant.} \\
    \hline 
  \end{tabular}
  \caption{Hallucinated \texttt{CN} generated by GPT-2 that is fluent and credible (according to Office for National Statistics, the Muslim population is just above 3M).}
 \label{tab:CN-wrong}
\end{table}

To this end, we explore methodologies to generate informative \texttt{CN}s using external knowledge. In particular, 
we propose an extension of knowledge-grounded generation approaches by adopting an intermediate step where we generate keyphrases for retrieving needed knowledge. So, we first train a counter narrative keyphrase generator, then the generated keyphrases are employed for selecting relevant knowledge sentences. Finally, pre-trained LMs are fine-tuned on the relevant knowledge sentences, together with the \texttt{HS} input, to produce knowledge-augmented \texttt{CN}s. Our extensive experiments on \texttt{CN} generation, including both automatic and expert evaluation, demonstrate that the presented approach produces more specific and tailored responses both for in-domain and zero-shot cross-domain configurations as compared to other approaches, such as standard LMs, that are simply fine-tuned for the task without the use of external knowledge.

As our main contribution, we show that: (i) external knowledge can boost informative \texttt{CN} generation, (ii)  keyphrase generation improves the quality of retrieved documents, (iii) silver knowledge is utilizable for the task when no gold knowledge is available, (iv) knowledge-bound models are advantageous while tackling zero-shot cross-domain generation especially if (v) using injection of knowledge in large pre-trained LMs.

\section{Related Work}
\label{sec:related_work}
In this section we review three main research topics that are relevant for fighting hatred online: (i) studies on \texttt{CN} effectiveness in hate countering, (ii) counter-argument generation and (iii) knowledge-guided generation.

\paragraph{Hate countering.} \label{sec:hate_countering_dataset}
Employing counter narratives has shown to be an effective strategy in hatred intervention on social media platforms. Studies have focused on identifying successful counter narratives \cite{benesch2016considerations, benesch2016}, evaluating their efficacy \cite{schieb2016governing,silverman2016impact,ernst2017hate,munger2017tweetment}, and analyzing counter speaker accounts characteristics \cite{mathew2018analyzing}. 
In particular, by analyzing conversations from Twitter, \citet{wright2017vectors} show that some arguments among strangers induce favorable changes in discourse and attitudes. 

\noindent\textbf{Counter-argument generation} shares similar objectives as \texttt{CN} generation, i.e., to produce the opposite or alternate stance of a statement, but the latter faces peculiar difficulties such as the absence in \texttt{HS} of explicit or well-structured `arguments' (e.g., ``\textit{Islam is a disease}") and the limited amount of data available for training. Studies usually focus on domains with large discussions, e.g., politics \cite{hua2018neural} and economy \cite{le2018dave}. The closest work to ours is counter-argument generation with external knowledge augmentation by \citet{hua-etal-2019-argument-generation}. Our approach differs from theirs in three aspects: (i) we explore generating queries to extract knowledge for grounding \texttt{CN} with, (ii) pre-trained generative models are utilized for leveraging the knowledge present, (iii) our approach requires less manipulation over knowledge. 

\noindent\textbf{Knowledge-guided generation.}
There is a growing interest in exploiting external knowledge to generate informative responses for applications such as dialog systems \cite{he2017learning, young2018augmenting} and question answering \cite{das2017question, saha2019complex}. Previous approaches inject knowledge through topic phrases \cite{fan2019strategies}, structured knowledge graphs \cite{zhou2018commonsense} and unstructured texts \cite{dinan2018wizard, hua-etal-2019-argument-generation}. 

\begin{figure*}[ht!]
\includegraphics[width=\textwidth,height=3.1cm]{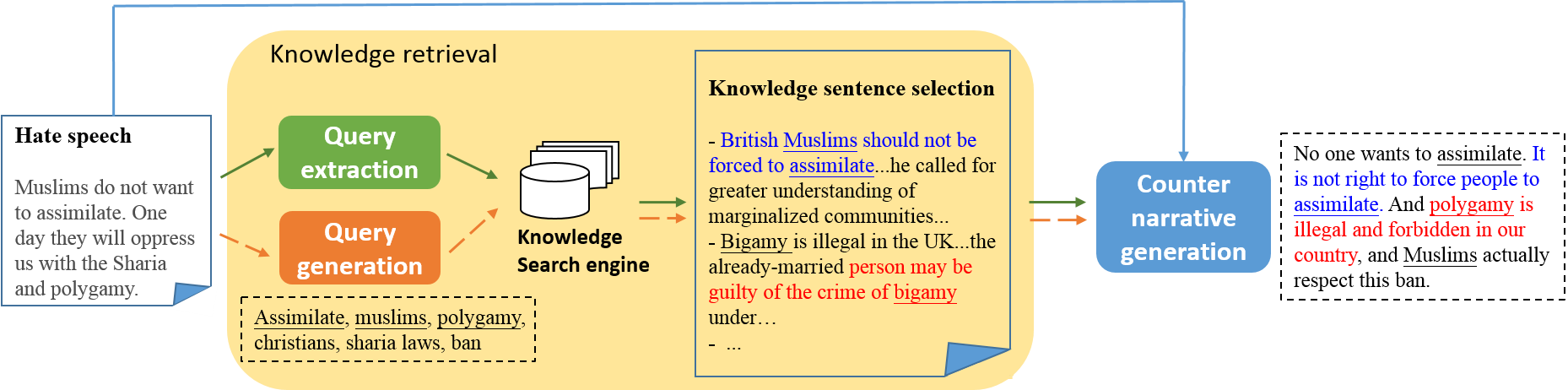}
\caption{Architecture of knowledge grounded generation with  extracted (green solid arrow) and generative (dotted arrow) queries (topical phrases) that are exploited to retrieve relevant knowledge. The knowledge sentences extracted together with input \texttt{HS} are fed to  \texttt{CN} generation. We give the example of generative approach.} 
\label{fig:architechture}
\end{figure*}

\section{HS-CN Dataset} To the best of our knowledge, there is no high-quality hate speech - counter narrative dataset available yet where \texttt{CN}s are explicitly paired with relevant knowledge. Since constructing such dataset with a decent-size would be too costly and out of the scope of the present paper\footnote{Obtaining access to a pool of trained NGO operators is very complicated, furthermore keeping track of their search activity and the material they used during \texttt{CN} production would require long and complex data collection sessions that might span several months.}, we resort to a ``reverse-engineering" strategy such that we automatically paired relevant knowledge with an already existing high quality \texttt{CN} dataset. We chose \texttt{CONAN} \cite{chung2019conan}, which is a dataset niche-sourced to expert NGO operators offering high quality \texttt{CN}s, and the best and most diverse material among the other \texttt{CN} datasets \cite{tekiroglu2020generating}. \texttt{CONAN} consists of 6645 English pairs of \texttt{HS}-\texttt{CN} including: 1288 original pairs, 2576 pairs where two paraphrases of the original \texttt{HS} are paired with the original \texttt{CN}, and 2781 translated pairs from French and Italian. The English data is split into 4069/1288/1288 samples for train/dev/test.

\section{Architecture} \label{sec:architecture}
Our architecture, illustrated in Figure \ref{fig:architechture}, consists of a knowledge retrieval module that retrieves sentence-level relevant knowledge, and a generation module that generates a counter narrative. Specifically, the knowledge retrieval module first prepares variants of a query $Q$ for a given hate speech \texttt{HS} using two strategies: query extraction ($Q_{hs}$) and automatic query generation ($Q_{gen}$). 
Then, the obtained queries are employed to search for relevant knowledge articles via a search engine. Finally, it uses a sentence selector to filter and rank the most relevant sentences as the relevant knowledge (\texttt{KN}) from the retrieved articles. For the counter narrative generation module, we fine-tuned several LMs that take a \texttt{HS} and the ranked knowledge sentences \texttt{KN} as input and output a corresponding counter narrative.

\section{Knowledge Retrieval Module}
The knowledge retrieval module in the architecture incorporates a knowledge repository, query construction sub-module, and a knowledge sentence selection sub-module. 

\subsection{Knowledge Repository} Previous approaches on introducing external knowledge for dialog generation have exploited unstructured and structured knowledge. Since no structured knowledge is available for the hate speech domain, we rely on unstructured textual knowledge in the format of articles, which allows for updating the knowledge repository easily. Considering that the proliferation of \texttt{HS} is also triggered with target-related events (e.g., terrorist attacks), being able to update the knowledge, such as news articles, would let us produce proper \texttt{CN}s that contain the latest statistics or evidence from the current events.

We include \texttt{Newsroom} \cite{grusky2018newsroom} and \texttt{WikiText-103} \cite{merity2016pointer} to our knowledge repository. \texttt{WikiText-103} is a large collection of 28,595 full Wikipedia articles covering over 103 million words. \texttt{Newsroom} consists of 1.3 million articles extracted from major news publications between 1998 and 2017, featuring over 6.9 million words. 

\subsection{Query Construction} \label{sec:guided_gen}
To construct comprehensive and proper queries to search for relevant knowledge for the data pairs, we applied two strategies: (i) query extraction and (ii) query generation.
In both strategies, the query is composed of keyphrases that can be defined as the important and topical phrases from a text \cite{turney2000learning}.   

\paragraph{Query extraction.}
We extracted keyphrases from \texttt{CONAN} dataset using Keyphrase Digger \cite{moretti2015digging}, a multilingual keyphrase extraction system that uses statistical measures and linguistic information, and is proven to be one of the best systems for unsupervised settings\footnote{Keyphrase Digger is a new implementation of KX \cite{pianta-tonelli-2010-kx} with several improvements, and it was ranked the best performing unsupervised system on task 5 of SemEval 2010 evaluation campaign \cite{moretti2015digging}.}. Following the knowledge retrieval strategy using input argument by \citet{hua-etal-2019-argument-generation} for counter argument generation, we first obtained the \texttt{HS} keyphrases to construct the initial query $Q_{hs}$. However, \texttt{HS}s from \texttt{CONAN} mostly contain hateful and simplistic phrases in comparison to the input arguments used by \citet{hua-etal-2019-argument-generation} that can be rich in content\footnote{e.g., an argument ``\textit{A universal basic income will help the labor market adapt to inevitable disruptions caused by advancements in automation and artificial intelligence.}" vs. \texttt{HS} ``\textit{Islam is a disease}".}. 
Therefore, in the \texttt{HS}-\texttt{CN} scenario, we hypothesize that the keyphrases from $Q_{hs}$ alone would not be sufficient for relevant knowledge search especially for mapping the knowledge onto training data.

To this end, we also extracted keyphrases from \texttt{CN} together with \texttt{HS} to increase the possibility that the retrieved knowledge sentences contain pieces of information found in the ground truth. Hence, the second query $Q_{hs \cup cn}$ contains \texttt{CN} keyphrases for the relevancy to the target \texttt{CN} and \texttt{HS} keyphrases for preserving the hate context. We investigated the effects of various keyphrase query configurations in terms of \texttt{HS} relevancy and $Q_{hs \cup cn}$ is proven to be the best configuration (See Appendix \ref{sec:keyphrase_configuration} for more details.). 

\paragraph{Query generation.} \label{sec:}
Since the best query configuration $Q_{hs \cup cn}$ cannot be available at test time, we need a way to obtain keyphrases that serve as \texttt{CN} cues for searching knowledge sentences during the \texttt{CN} generation. To this end, we built a query generation model that takes \texttt{HS} as input and outputs a comma-separated list of \texttt{CN} keyphrases, which is then used as $Q_{gen}$. Our aim is to obtain an approximation of $Q_{hs \cup cn}$ via $Q_{hs \cup gen}$ at the test time.

The model is trained using Transformer \cite{vaswani2017attention} architecture as it has obtained state-of-the-art performances for generation tasks \cite{dinan2018wizard, ghazvininejad2018knowledge}. For the training data, we used \texttt{CONAN} dataset and discarded the \texttt{CN}s that are 
less than 10 words, since they are usually generic, poor in terms of argumentative content and cannot provide a meaningful search (e.g., ``\textit{No they are not - prove this?}'', ``\textit{What does that even mean?}'', ``\textit{Any evidence?}'').  
Accordingly, we kept 4038/1257/1257 instances for train/dev/test set. The train set includes the pairs marked as original in the dataset, and all translated pairs from French and Italian; the dev set consists of one paraphrase of each original \texttt{HS} and its \texttt{CN}s; and the test set contains the rest of the paraphrased \texttt{HS}s.  
The training inputs are represented as $\;HS\; [HS\_end\_token]\; KP \; [KP\_end\_token]$, where $KP$ is the list of keyphrases extracted from the gold \texttt{CN}.

The model has been trained following the configuration of the base model in \citep{vaswani2017attention}: with 6 transformer layers, 8 transformer heads, embedding size of 512, hidden size of 2048, dropout rate of 0.1, batch size of 64 for 100 epochs. The training time lasted around 7 hours. All experiments in this paper have been conducted on a Nvidia Tesla V100 GPU. For decoding, we used nucleus sampling \cite{holtzman2019curious} with a p value of 0.9.

We report keyphrase generation results in terms of BLEU \cite{papineni-etal-2002-bleu} and ROUGE \cite{lin-2004-rouge} against the keyphrases extracted from gold test \texttt{CNs}.
We obtained a score of 0.162 for BLEU-2 and a score of 0.353 for ROUGE-L. Although both scores can be considered as low, this is due to the open ended nature of the set of possible \texttt{CN} keyphrases for a given \texttt{HS}. Example queries for a single pair, extracted from its \texttt{HS}, its \texttt{CN}, and generated with the keyphrase generation model are shown in Table \ref{tab:query}.

\begin{table*}[ht!]
\small
\centering
\begin{tabularx}{\textwidth}{c|l|p{2.5cm}|X} 
    \hline
    \texttt{HS} & \texttt{CN} & Query & Knowledge sentences (\texttt{KN}) \\
    \hline
    \multirow{3}{3em}{Islam is a disease.} & \multirow{3}{7em}{Like Christianity or any other, islam is a religion of tolerance. Disease does not discriminate on religious grounds. 
    } & $Q_{hs}$: islam, disease &  (i) Do Muslims want to heal from the disease...? (ii) Being infected by religious extremism is like being infected by a disease...  \\ \cline{3-4}
    & & $Q_{gen}$: islamic law, god, christians & (i) Islamic law is to create an environment...submission to God. (ii) Certain areas of the Muslim world have always been home to large populations of Christians...  
\\ \cline{3-4}
    & & $Q_{cn}$: tolerance, christianity, discriminating & (i) Islam is a 1400-year-old religion that preaches tolerance...like Christianity and... (ii) Disease does not discriminate...on religious grounds... \\
    \hline
\end{tabularx}
\caption{Examples of \texttt{KN} retrieved using queries extracted from \texttt{HS} ($Q_{hs}$), generated ($Q_{gen}$) and created from both \texttt{HS} and \texttt{CN} keyphrases ($Q_{cn}$).}
\label{tab:query}
\end{table*}

\subsection{Knowledge Sentence Selection}
\label{subsec:sentence-selection}
We use \texttt{Solr}\footnote{\url{https://lucene.apache.org/solr/}} to index the articles and retrieve those relevant to a given query based on the similarity between the articles and the query using BM25 \cite{robertson1995okapi}. 
Once the queries have been obtained either through extraction or generation, they are presented to \texttt{Solr} for retrieving the 25 top-ranked articles.
Next, we used spaCy sentence segmentation\footnote{\url{https://spacy.io/universe/project/spacy-sentence-segmenter}} to split an article into sentences. Similar to \citet{zhang2019pegasus}, given a query $Q$ we score each sentence $x_i$ in the set of articles $D$ independently, using ROUGEL-F1 \cite{lin-2004-rouge} as in Equation~\ref{equation:text_generation}. 
\begin{equation}
\label{equation:text_generation}
   s_i = rouge(x_i, Q), \forall i \in D
\end{equation}

In the final step, we distilled the knowledge by keeping the top 5 knowledge sentences that have the highest scores among 25 top-ranked articles. Instead of a more stringent filtering, such setting has been applied to grant a better variety of source articles and corresponding distilled sentences. We refer to such automatically associated sentences as ``silver knowledge".

\section{Counter Narrative Generation Module}
Large pretrained LMs require less amount of high-quality data to be fine-tuned  on downstream tasks while providing strong performances and they already store large amount of factual and commonsense knowledge from their training data \cite{petroni2019language}. To this respect, we built the following models: (1) \textbf{GPT-2$_{KN}$}, obtained by fine-tuning GPT-2 on \texttt{CONAN} data paired with \texttt{KN}; 
(2) \textbf{GPT-2$_{KN,MT}$}, by fine-tuning GPT-2$_{KN}$ in a multi-task learning fashion for learning to distinguish \texttt{CNs} from \texttt{HS} as next utterances; (3) \textbf{XNLG} \cite{chi2019cross} for its ability to copy information to the output (in our case the retrieved \texttt{KN} to be copied to the \texttt{CN}). 
We expect all three models to attend over the \texttt{HS} and retrieve \texttt{KN} and look for the relevant snippets to be recovered while generating a \texttt{CN}.

\subsection{Models}
\label{subsec:models}
The training \texttt{HS}-\texttt{CN} pairs are represented as $\;HS\; [HS\_end\_token]\; KN \; [KN\_end\_token]$ $\; CN \; [CN\_end\_token]$. Each model is trained with $Q_{hs \cup cn}$ and then tested on  $Q_{hs}$, $Q_{gen}$, and $Q_{hs \cup gen}$. We also tested the models with $Q_{hs \cup cn}$ to define an oracle scenario with an upperbound performance when the data can only be paired with silver knowledge.

\paragraph{GPT-2$_{KN}$.} We fine-tuned the GPT-2\footnote{\url{https://github.com/huggingface/transformers}} medium model for 3 epochs with a batch size of 2048 tokens. We used Adam optimizer with a learning rate of 5e-5. At inference time, responses were generated employing nucleus sampling with a p value of 0.9, conditioned on \texttt{HS}s and corresponding \texttt{KN}.

\paragraph{GPT-2$_{KN,MT}$.} 
Since we noticed that GPT-2 occasionally produces responses that contain fragments of abusive language, we combined the language modeling objective with a next-sentence prediction objective for fine-tuning GPT-2 in a multi-task setting, inspired by \citet{DBLP:journals/corr/abs-1901-08149}. Next-sentence prediction adopts a linear classification layer added to the last layer of the transformer language model and then applies a cross-entropy loss to classify a proper next response to the input \texttt{HS} from 2 distractors randomly selected from \texttt{HS}.
We used Adam optimizer with a learning rate of 5e-5 and empirically fine-tuned it for 1 epoch and the same sampling strategy as GPT-2$_{KN}$ has been applied. 

\paragraph{XNLG} is a pre-trained Transformer-based language model trained on Wikipedia dumps with two relevant objectives for our task: to obtain contextual representations and to recover a given input. We fine-tuned XNLG\footnote{\url{https://github.com/CZWin32768/XNLG}} for generating counter narratives on all layers with a batch size of 10 for 100 epochs. We used Adam optimizer with a learning rate of 1e-4. 
We tokenized and removed accent from the entire dataset and applied the same BPE codes used by \citet{chi2019cross}. For \texttt{KN} and \texttt{CN} we kept the first 256 tokens, while setting the \texttt{HS} to 70 tokens, which is the maximum length of hate speech in the dataset. 
We experimented with various decoding methods and adopted beam search with a beam-width of 3 for the best performing setting (details in Appendix \ref{sec:decoding_methods}).

\paragraph{Baselines} \label{sec:unguided_gen} used for comparison are: (1) non-pretrained \textbf{Transformer} without knowledge using the same hyper-parameters as keyphrase generation model; (2) \textbf{GPT-2} without knowledge following the same configuration as GPT-2$_{KN}$; (3) \textbf{Candela} \cite{hua-etal-2019-argument-generation}, an LSTM-based state-of-the-art knowledge-driven architecture for argument generation. 
Since \texttt{CONAN} is relatively small, we hypothesize that a pre-training procedure\footnote{We also trained Candela from scratch on \texttt{CONAN} but decided not to proceed with this setting for the poor performance.} on data from a similar task (argument generation) can be beneficial for generalization and porting knowledge. Thus, we first pre-trained Candela architecture on argument generation dataset \cite{hua-etal-2019-argument-generation}, following the configuration described in the paper. We then fine-tuned the model for 20 epochs on \texttt{CONAN} with \texttt{KN} using $Q_{hs}$ as it is done in the original setting of Candela.

\begin{table*}[ht!]
\small
  \centering
  \begin{tabular}{l r r r r r r  r r r }
    \hline
    && &&&& &\multicolumn{3}{c}{\texttt{KN} overlap (ngram)}  \\
    Models &  Nov. & RR & B-2 & R-L &  \#Word & \#Sent. & 1 & 2 & 3  \\
    \hline 
    \textit{without knowledge} \\
    \, TRF & 0.467 & 7.72 & 0.082 & 0.094 &  21.47 & 1.70  & -& -& -\\
    \, GPT-2 & 0.688 & 9.04 & 0.045 & 0.100 & 15.95 & 1.35 & -& -& - \\
    \, Train$_{cn}$ & - & 3.91 & - & - & 21.79 & 1.87  & 0.307 & 0.054 & 0.016 \\
    \hline
    \hline
    \textit{with knowledge} \\
    \, Candela ($Q_{hs}$) & 0.692 & 21.87 & 0.040 & 0.098 & 23.85 & 2.47  & 0.173 & 0.008 & 0.001  \\
    \hline
    \, \textbf{GPT-2$_{KN}$} &&&&&& & & & \\
    \, \, w/ $Q_{hs}$ & 0.723 & 8.13 & 0.082 & 0.094 & 15.60 & 1.32 & 0.258 & 0.023 & 0.008 \\ 
    \, \, w/ $Q_{gen}$ & 0.728 & 7.48 & 0.067 & 0.091 & 12.75 & 1.17  & 0.260 & 0.050 & 0.019 \\
    \, \, w/ $Q_{hs \cup gen}$ & 0.735 & 6.30 & 0.085 & 0.103 & 15.35 & 1.59  & 0.358 & 0.068 & 0.024  \\
    \, \, w/ $Q_{hs \cup cn}$ & 0.727 & 7.17 & \textbf{0.166} & 0.110 & 13.10 & 1.16  & 0.282 & 0.058 & 0.022  \\
    \hline    
    \, \textbf{GPT-2$_{KN,MT}$} &&&&&& & & & \\
    \, \, w/ $Q_{hs}$ & 0.744 & 11.69 & 0.050 & 0.090 & 13.35 & 1.17 & 0.269 & 0.049 & 0.017 \\
    \, \, w/ $Q_{gen}$ & 0.731 & 10.37 & 0.052 & 0.092 & 13.34 & 1.14  & 0.253 & 0.044 & 0.017 \\
    \, \, w/ $Q_{hs \cup gen}$ & 0.747 & 7.59 & 0.091 & 0.090 & 16.91 & 1.26  & 0.269 & 0.033 & 0.009 \\
    \, \, w/ $Q_{hs \cup cn}$ & 0.731 & 9.56 & 0.048 & 0.107 & 13.05 & 1.13  & 0.276 & 0.057 & 0.023  \\
    \hline
    \, \textbf{XNLG} &&&&&&  & & & \\
    \, \, w/ $Q_{hs}$ & \textbf{0.824} & 14.42 & 0.073 & 0.084 &  55.51 & 3.71 & 0.841 & 0.650 & 0.558  \\ 
    \, \, w/ $Q_{gen}$ &  0.819 & 6.88 & 0.097 & 0.084 &  55.64 & 3.64  & 0.849 & 0.656 & 0.558  \\
    \, \, w/ $Q_{hs \cup gen}$ &  0.812 & 6.98 & 0.074 & 0.089 &  57.58 & 3.00  & 0.828 & 0.579 & 0.475 \\
    \, \, w/ $Q_{hs \cup cn}$ &  0.819 & \textbf{5.69} & 0.076 & \textbf{0.116} &  55.69 & 3.42 & 0.840 & 0.631 & 0.529  \\
    \hline
  \end{tabular}
  \caption{Results of \texttt{CN} generation with silver knowledge. We report novelty (Nov.), RR, BLEU-2 (B-2), ROUGE-L (R-L), \texttt{KN} overlap with generation and the average amount of words and sentences per generation.
  }\label{tab:guided_generation}
\end{table*}

\subsection{Results for the Silver Knowledge Test Set}
We report BLEU-2 (B-2) and ROUGE-L (R-L) scores for all proposed models and baselines in Table~\ref{tab:guided_generation} on the test split of \texttt{CONAN} that we automatically paired with silver knowledge using various queries. We also measure the capability of each model to produce \textit{novel} responses with respect to the training data by Jaccard similarity \cite{wang-2018-sketching}, and \textit{diverse} responses for the given input by repetition rate  (RR)~\cite{cettolo2014repetition}.

Among our models, GPT-2$_{KN}$ yields the highest B-2 and XNLG the highest novelty, diversity, and R-L. 
The notably improved novelty achieved by knowledge-grounded models indicates the benefits of adding knowledge on producing \texttt{CN}s, in comparison to the baselines - particularly Transformer \texttt{TRF}. 
On the other hand, the quantitative performance of XNLG does not reflect its true performance in terms of quality. A quick glance at the output \texttt{CN}s showed that XNLG model copies almost everything from \texttt{KN} to the output instead of a proper \texttt{CN} generation, increasing the novelty and diversity scores. The issue can easily be observed from the average numbers of words and sentences in the XNLG output in comparison to the outputs of the other models presented in Table~\ref{tab:guided_generation}.
GPT-2$_{KN,MT}$ falls behind among our models in terms of RR, B-2, and R-L, still providing a competitive novelty. Regarding Candela, while it obtained similar performances to our models in terms of R-L and B-2, the generation is repetitive and less novel. 

As for the testing with different query types, $Q_{hs \cup gen}$ induces more novel responses than $Q_{hs \cup cn}$ and $Q_{hs}$. 
While XNLG yields the highest novelty with $Q_{hs}$ (0.824), it can be explained again with the problem of copying the whole \texttt{KN}, which is more varied due to the less restrictive search using only \texttt{HS}. 

The oracle query $Q_{hs \cup cn}$, in which we deliberately provide the best knowledge possible through the keyphrases containing also from the gold \texttt{CN}, yields the best R-L scores among the query variations of knowledge-grounded models. Among all the models, $Q_{hs \cup cn}$ also leads to the best B-2 through GPT-2$_{KN}$, and the best R-L through XNLG, as we have anticipated. 
Finally, $Q_{hs \cup gen}$ outperforms $Q_{hs}$ and $Q_{gen}$ over most metrics, hinting at the advantages of using generated queries together with hate context for silver-knowledge retrieval. 

We have also conducted complementary experiments by taking into consideration the design choices and the various phenomena in our study. Since in our test set, in line with \texttt{CONAN}, a \texttt{HS} can be paired with more than one \texttt{CN}, $Q_{hs}$ would retrieve the same \texttt{KN} for all the target \texttt{CN}s of the same input \texttt{HS}. Contrarily, we obtain a different set of \texttt{KN} using queries $Q_{gen}$, $Q_{hs \cup gen}$ and $Q_{hs \cup cn}$ for each target \texttt{CN}. Therefore, we also report an evaluation on unique \texttt{HS}-\texttt{CN} pairs, where a single target \texttt{CN} has been randomly chosen for each \texttt{HS}, among all query types in Appendix \ref{sec:unique_hs}.
Finally, to simulate Candela configuration (that uses only $Q_{hs}$) also with the other models, we run an additional set of experiments where we used $Q_{hs}$ for retrieving the knowledge for training samples. The results are reported in Appendix \ref{sec:Qcn_generation}.

\begin{table*}[t]
\small
\centering
\begin{tabular}{lrcrrrrcrr}
\hline
&\multicolumn{4}{c}{in-target} & &\multicolumn{4}{c}{cross-target} \\\cline{2-5}\cline{7-10}
& Nov. & RR &B-2 & R-L & & Nov. & RR &B-2 & R-L   \\ \cline{2-10}
TRF & 0.30 & 7.57 & 0.014 & 0.10 && 0.46 & 8.62 & 0.015 & 0.08 \\
GPT-2 & 0.72 & 8.53 & 0.020 & 0.11 && 0.72 & 8.01 & 0.022 & 0.09 \\
Candela & 0.69 & 19.31 & 0.072 & 0.10 && 0.70 & 22.22 & 0.022 & 0.09 \\
GPT-2$_{KN}$ & 0.71 & \textbf{6.85} & 0.201 & 0.19 && 0.75 & \textbf{6.33} & 0.041 & 0.19 \\
GPT-2$_{KN,MT}$ & \textbf{0.85} & 11.55 & 0.066 & 0.12 && \textbf{0.86} & 10.38 & 0.022 & 0.11  \\
XNLG  & 0.83 & 6.94 & \textbf{0.256} & \textbf{0.33} && 0.84 & 8.15 & \textbf{0.291} & \textbf{0.35} \\
\hline
\end{tabular}
\caption{Results of \texttt{CN} generation with gold knowledge in-target and cross-target test sets. }
\label{tab:optimum_test}
\end{table*}

\subsection{Results for the Gold Knowledge Test Set}
To isolate the effect of the knowledge retrieval strategies from the knowledge-grounded generation performances, we conducted a second evaluation on a newly crafted test set paired with gold standard knowledge. In this evaluation, in addition to stereotypical islamophobic in-domain (i.e. in-target) scenario, we also explore the effect of knowledge infusion on cross-domain (i.e. cross-target) \texttt{CN} generation under zero-shot setting. We hypothesize that having a system trained to make use of substandard silver knowledge to generate proper \texttt{CN}s for a given context, could be robust to cross-domain zero-shot conditions. Therefore, we organized a data collection session with an expert operator in writing \texttt{CN}s. In this session, 50 islamophobic \texttt{HS}s randomly sampled from \texttt{CONAN} and 144 new cross-target \texttt{HS}s (covering misogyny, antisemitism, racism, and homophobia) are provided along with the knowledge retrieved by $Q_{hs \cup cn}$ queries. The expert is tasked with composing a suitable \texttt{CN} using the corresponding knowledge as much as possible. Thus, we could obtain a gold test set\footnote{We release the gold test set at \url{https://github.com/marcoguerini/CONAN}.}. in which the input knowledge can certainly be found in the \texttt{CN}s. 

We tested all models with gold knowledge in-domain and cross-domain test cases. Results are given in Table~\ref{tab:optimum_test}. For in-domain scenario, as we have anticipated, knowledge grounded models yield better performances in B-2 and R-L in comparison to the silver knowledge test setting. Especially with the striking jump in the performance of GPT-2$_{KN}$, we can confirm the proper infusion of the given knowledge to the generated \texttt{CN}s. As for cross-domain tests, GPT-2$_{KN}$ still yields better performance than baselines while the performance for all models (except for XNLG) drops due to unseen events during training. All GPT-2 variations present better diversity performances on the cross-domain setting as compared to both in-domain and silver-knowledge settings. 
Regardless of domains, XNLG yields fallaciously high scores due to its extensive copying. 
A cross-target generation from the models can be seen in Table~\ref{tab:CN-cross-domain-examples}. More examples in-/cross-domain generations from all the models are provided in Appendix \ref{sec:sample_generation}. 

\paragraph{Human evaluation.}
We further resort to human evaluation to assess the final generation quality of each model. For this reason we perform human evaluation of generation using gold knowledge, to rule out the effect of possible noise in the knowledge that may result from the retrieval process. 

Our models were evaluated by 3 expert operators from the NGO Stop Hate UK. The annotators are already experienced, and specifically trained, in reading hateful content and writing \texttt{CN}s for online hate countering\footnote{The compensation for annotation work met with EU regulations.}. 
The annotators are instructed to assess all generated pairs in gold knowledge test sets in terms of \emph{suitableness} to the \texttt{HS}, \emph{informativeness}, and \emph{intra-coherence} of \texttt{CN} regardless of \texttt{HS}.
Each score is on a scale of 1 (the least) to 5 (the most). To avoid possible bias and hints towards models, we normalized the pairs (e.g., lowercase and space between words and punctuation) and divided them into 3 partitions of randomized files for experts (See Appendix~\ref{sec:evaluation_instruction} for annotation instruction). 
Each expert was given 388 pairs, resulting in a total of 1164 pairs for evaluation. To avoid excessive workload annotators were allowed to complete the task over multiple sessions at their preference. 

Results are reported in Table \ref{tab:optimum_human_evaluation}. We also computed l's Tau-b \cite{10.2307/2332226} to measure the annotators' agreement towards the model ranking for each aspect. The high correlations indicate a strong concordance among the annotators (threshold tau-b $> 0.35$). Regardless of domains, annotators consider XNLG generations as the most informative and GPT-2$_{KN}$ generations as the most suitable. 
TRF yields a reasonable suitableness and coherence since it tends to memorize the training \texttt{CN}s, almost behaving like a retrieval system on human responses. However, such behavior can be fatal in cross-domain settings. Candela fails to generate suitable cross-domain \texttt{CN}s despite preserving the intra-\texttt{CN} coherence. 
While GPT-2 and GPT-2$_{KN}$ generations are found almost equally coherent, the lower suitableness and informativeness of GPT-2 output (2.26 and 1.92) for cross-domain as compared to GPT-2$_{KN}$ (2.51 and 2.29) encourages the grounding \texttt{CN}s in knowledge. 

\begin{table}[t]
\small
\resizebox{\columnwidth}{!}{
\begin{tabular}{l ccc l ccc }
\hline
&\multicolumn{3}{c}{in-domain} & &\multicolumn{3}{c}{cross-domain} \\\cline{2-4}\cline{6-8}
& suit. & info. & cohe. && suit. & info. & cohe.   \\ \cline{2-8}
TRF       &  2.65 & 2.25 & 3.39 && 1.47 & 2.09 & 3.45 \\
GPT-2   & 2.67 & 2.16 & 4.10 && 2.26 & 1.92 & \textbf{4.24} \\
Candela   & 2.41 & 2.25 & 3.14 && 1.42 & 2.09 & 3.40 \\
GPT-2$_{KN}$   & \textbf{3.02} & 2.35 & \textbf{4.33} && \textbf{2.51} & 2.29 & 4.21 \\
GPT-2$_{KN,MT}$ & 1.76 & 1.65 & 3.73 && 2.03 & 1.76 & 3.88 \\
XNLG      & 1.43 & \textbf{3.88} & 2.12 && 1.88 & \textbf{4.10} & 2.79 \\
\hline
Kendall's tau-b & 0.82 & 0.69 & 0.82 && 0.51 & 0.91 & 0.73 \\
\hline
\end{tabular}
}
\caption{Human evaluation results of \texttt{CN} generation.}
\label{tab:optimum_human_evaluation}
\end{table}

\section{Discussion}
Our findings suggest that a large pre-trained LM with knowledge injection is preferred to alleviate the demand for gold data and improves in-/cross-domain generations. GPT-2$_{KN}$ outperforming GPT-2, which becomes more clear with every increase in the quality of provided \texttt{KN} (i.e., from silver $Q_{gen}$ to silver $Q_{hs \cup gen}$ to gold in-domain), indicates that our model trained on silver knowledge learns to grasp relevant \texttt{KN} to the given \texttt{HS} instead of random copying. 
As for GPT-2$_{KN,MT}$, since the model is trained to distinguish \texttt{CN} from \texttt{HS}, certain typical counter responses are prone to appear. Through a manual analysis, we confirmed that its responses tend to be simpler (e.g., negation: ``\textit{terrorism is \textbf{not} a religion.}'', ``\textit{It's \textbf{not} true...}'') than the responses of other GPT-2 variant models. Future work can explore alternative multitask learning frameworks to improve the generation. 

From a practical perspective, we can conclude that fine-tuning a pretrained language model for conditional generation, utilizing silver-knowledge sentences retrieved through a query generation model is the most promising approach for knowledge-bound \texttt{CN} generation. In fact, (i) among various models we tried, GPT-2 with knowledge infusion is shown to produce the most informative and suitable output. (ii) At deployment time, using input context together with generated keyphrases becomes prominent with more novel and diverse \texttt{CN}s and (iii) for new hate-targets in a zero-shot scenario, the generation model can produce coherent and suitable \texttt{CN}s without retraining if a reasonable set of input knowledge is provided. Moreover, this approach can be applied to other scenarios, such as dialogue response generation, since it does not require any extra data collection or annotation.

\begin{table}[ht!]
  \small
\resizebox{\columnwidth}{!}{
  \begin{tabular}{p{\columnwidth}}
    \hline
\textbf{HS:} Jews are different from non-Jews not because of their beliefs, but because of their physical nature. “Jewishness” and malevolence is inherited through blood.\\
\hline
\textbf{Human:} \textcolor{blue}{Today one of the biggest concerns is the reinvention of clinical concepts of race, based on inaccurate generalizations}. The notion that the superiority of the white race was part of the natural order was deviously reinforced by the rise of modern "scientific" racism in the late 19th century. Despite the strong objections of many of our contemporaries, this idea of a "jewish race" is still going around. \\
\hline
\textbf{GPT-2:} Muslims are not different from non-Muslims because of their faith. \\
\hline
\textbf{GPT-2$_{KN}$:} I think that the \textcolor{blue}{notion of race} is a \textcolor{blue}{generalization based on} ignorance. \\
\hline
\textbf{GPT-2$_{KN,MT}$:} All that is wrong is that we cannot accept that violence is necessary. \\
\hline
\textbf{XNLG:} \textcolor{blue}{today , one of the biggest concerns is the reinvention of clinical concepts of race , based on inaccurate generalizations} about the prejudice to certain physical characteristics and civilizations . despite the strong objections of many contemporary contemporaries , the notion of fixed " race " - packages of physical and behavioral characteristics \\
    \hline 
  \end{tabular}
  }
  \caption{Samples of cross-domain generation.}
 \label{tab:CN-cross-domain-examples}
\end{table}

\section{Conclusion}
\label{sec:conclusion}

Online hate speech intervention is a challenging problem and research on counter narrative generation is still in its infancy. In this work, we have proposed methods for improving counter narrative generation to fight hatred online, incorporating external knowledge retrieved through extracted and generated keyphrases. To this end, a dataset of hate-speech/counter-narrative pairs was augmented with relevant knowledge to train systems that are able to produce suitable and informative arguments. 
Our experiments on in-/cross- domain generation indicate that the generated responses can meet these desiderata. As future work we plan to test other query generation approaches (e.g., exploit human-crafted queries in an interactive setup or the use of pre-trained LMs) to improve knowledge selection and to test other architectures for the final generation step. 

\section*{Acknowledgments}

This work was partly supported by the HATEMETER project within the EU Rights, Equality and Citizenship Programme 2014-2020. We are deeply grateful to Stop Hate UK and its volunteers for their help and effort in evaluating the output of our systems.

\section*{Ethical Considerations}
While we believe that counter narratives are a better tool than content moderation in fighting hate speech (e.g. they do not hinder freedom of speech), still the automatic generation of CNs should be taken with care. Since this work aims at presenting a methodology for knowledge-bound counter narrative production through neural approach, several ethical consequences should be considered.

First, neural models may still produce substandard counter narratives containing abusive language or negative content. To mitigate this issue, possible solutions include (1) integrating in the pipeline a classifier or a human reviewer for validation and possible post-editing \cite{tekiroglu2020generating}, (2) detoxification techniques for controllable generation methods \cite{gehman-etal-2020-realtoxicityprompts}, and (3) discarding undesirable content from the corpora used for training \cite{JMLR:v21:20-074}, even if the appropriate criteria for such purpose are still investigated.

Second, while our approach reduces the risks of content hallucination, an additional step, where the accuracy of the generated text is checked against the provided knowledge \cite{nie-etal-2019-simple,dusek-kasner-2020-evaluating}, would provide further robustness to the system. 

Third, natural language generation models may still induce unintended social biases. This issue can be moderated by measuring/promoting fairness in models and data employed \cite{blodgett-etal-2020-language}, and designing bias triggers \cite{sheng-etal-2020-towards} or regularization methods \cite{bordia-bowman-2019-identifying, 10.1145/3097983.3098095} for controllable bias.

To sum up, while some additional automated techniques may help in maintaining generation quality, human evaluation should always be considered as the foremost solution, at least for delicate tasks such as `real' hate countering on social media platforms.  For this reason we advocate that generation systems should be used as a suggestion tool for NGO operators, to make their countering work more effective.  In this way there is always a “human moderator” taking the final decision \cite{chung2019conan}.

\bibliography{anthology,acl2021}
\bibliographystyle{acl_natbib}
\clearpage
\appendix

\section{Appendices}

\subsection{Analysis on Keyphrase Extraction Configurations} \label{sec:keyphrase_configuration}
We conducted a preliminary manual analysis to investigate the effects of various keyphrase extraction configurations. 
We randomly sampled 48 hate speech and counter-narrative pairs from \texttt{CONAN} dataset and extracted the keyphrases. Then, we retrieved the \texttt{KN} (see Section \ref{subsec:sentence-selection}) with the queries $Q_{hs}$ and $Q_{hs \cup cn}$. On the other hand, we also wanted to inspect the condition with the keyphrases only from \texttt{CN}, i.e., $Q_{cn}$. For each sample and condition, annotators have assigned a score for relevance to the hate speech in the scale of 1 to 5; 1 meaning no-relevance, and 5 perfect relevance. As a result, we have noticed that $Q_{cn}$ is the worst condition, i.e., non-optimal, having an average score of $2.30$. The analysis shows that it causes the loss of context related to \texttt{HS}, bringing information mostly from whole another topic. For instance, especially when \texttt{CNs} are rather generic, often, no lexical hint can be found related to the topic of Islamophobia (e.g., “\textit{Do you have any proof?}”). Indeed, $Q_{hs}$ provides an apparently better average score ($3.46$) since it provides a better context to search for. However, as expected, the best score ($3.77$) has been obtained through $Q_{hs \cup cn}$, i.e, optimal, verifying our hypothesis of utilizing both \texttt{HS} and \texttt{CN} keyphrases for training.

\subsection{Preliminary Analysis on Decoding Methods for XNLG} \label{sec:decoding_methods}
To find a suitable decoding method for our task, we generated \texttt{CN}s with 3 candidate settings: beam search with a beam-width of 3 and top-k sampling with a k value of 8 and 10. For each setting we utilized \texttt{KN} retrieved with both non-optimal ($Q_{cn}$) and optimal ($Q_{hs \cup cn}$) queries. 
Then we sampled 120 \texttt{HS}-\texttt{CN} pairs and served them to three experts in \texttt{CN} writing for evaluating the generation on a scale of 1 (the worst) to 5 (the best) in terms of suitableness and informativeness. Suitableness measures if the generated \texttt{CN} is relevant to the \texttt{HS} and informativeness evaluates the amount of information (e.g. statistics and facts) enclosed in the \texttt{CN}.

The results reported in Table \ref{tab:decoding_evaluation} reveal a clear difference between beam search and top-k sampling regardless of \texttt{KN} being optimal or non-optimal. In a manual investigation, we observed that the generation using both beam search and top-k sampling generally can copy some pieces of information from the given \texttt{KN}, while top-k seems to replace part of the text with slightly relevant and uncommon words. 
Hence, copying the right knowledge pieces through the decoding strategy is a key factor instead of diverging from the knowledge solely for the sake of lexical diversity. 
Therefore, based on the results, we adopt beam search with a beam-width of 3, which is shown to be the most suitable and informative, for decoding method in our experiments. 

\begin{table}[ht!]
  \centering
  \begin{tabular}{l r r }
    \hline
    Decoding methods & Suit. & Info. \\
    \hline 
    \textit{Non-optimal knowl.} & &\\
    Beam-3 & 1.950 & 2.325 \\
    Topk-8 & 1.275 & 1.775 \\
    Topk-10 & 1.625 & 2.100 \\
    \hline 
    \textit{Optimal knowl.} & &  \\
    Beam-3 & 2.325 & 2.450 \\
    Topk-8 & 1.975 & 2.175 \\
    Topk-10 & 2.050 & 2.325\\
    \hline
  \end{tabular}
  \caption{Human evaluation on \texttt{CN} generation using various decoding methods.}\label{tab:decoding_evaluation}
\end{table}

\subsection{CN Generation with $Q_{hs}$
} \label{sec:Qcn_generation}

In this section we report the \texttt{CN} generation results of our knowledge-bound models trained and tested with $Q_{hs}$. We applied the same hyperparameter configurations as the models trained with $Q_{hs \cup cn}$ described in Section \ref{subsec:models}.

The results are given in Table \ref{tab:Qcn_generation}. In contrast to the baselines (i.e., models without knowledge and Candela), all models obtained higher novelty with $Q_{hs}$. The repetition rate, on the other hand, is not improved since the models exploit the same knowledge for multiple test samples due to the repeated \texttt{HS}s with different \texttt{CN}s in the test set. 

We also observed that for GPT-2$_{KN}$ and GPT-2$_{KN,MT}$ the generation with $Q_{hs}$ is more repetitive and less novel compared to the generation applying queries $Q_{hs \cup gen}$ (as shown in Table \ref{tab:guided_generation}). This result demonstrates the viability and necessity of using generated queries, as potential \texttt{CN} prompts, along with \texttt{HS} context.

\begin{table*}[t]
\small
  \centering
  \begin{tabular}{l r r  r r r r  r r r}
    \hline
    && &&&& &\multicolumn{3}{c}{\texttt{KN} overlap (ngram)}  \\
    Models &  Nov. & RR & B-2 & R-L &  \#Word & \#Sent. & 1 & 2 & 3  \\
    \hline
    GPT-2$_{KN}$ & 0.700 & 8.66 & 0.082 & 0.098 & 14.87 & 1.32  & 0.235 & 0.019 & 0.003 \\
    GPT-2$_{KN,MT}$ & 0.730 & 11.92 & 0.084 & 0.090 & 13.25 & 1.13  & 0.233 & 0.025 & 0.007  \\
    XNLG & 0.824 & 16.46 & 0.073 & 0.084 &  55.51 & 3.71 & 0.841 & 0.650 & 0.558\\
    \hline
  \end{tabular}
  \caption{Results of \texttt{CN} generation applying $Q_{hs}$.}
  \label{tab:Qcn_generation}
\end{table*} 

\subsection{Unique HS Test Set Analysis} \label{sec:unique_hs}
Concerning that one \texttt{HS} can be paired with different \texttt{CN}s in the test, 
we further conducted an evaluation on a unique set by keeping each unique \texttt{HS} and one randomly selected \texttt{CN} among its \texttt{CN}s. The unique \texttt{HS} set lets us perform a fairer comparison among query configurations especially for $Q_{hs}$ with models employing beam search (i.e. XNLG and Candela). The results are given in Table \ref{tab:unique_test}. 

For XNLG and GPT-2$_{KN,MT}$, we observed an increase in the novelty and diversity with $Q_{hs}$ and $Q_{gen}$ on the unique \texttt{HS} set over the whole test set. As for GPT-2$_{KN}$, while diversity improves for all query configurations, we did not observe an increase in novelty through $Q_{hs}$. For Candela, while novelty also increases, the diversity does not improve. 

\begin{table}
\small
  \centering
  \begin{tabular}{l r r }
    \hline
    &&   \\
    Models &  Nov. (W/U) & RR (W/U)  \\
    \hline 
    \textit{without knowledge} \\
    \, TRF & 0.467/0.457 & 7.72/6.97 \\
    \, GPT-2 & 0.688/0.675 & 9.04/7.79 \\
    \, Train$_{cn}$ & - & 3.91/3.31  \\
    \hline
    \hline
    \textit{with knowledge} \\
    \, Candela ($Q_{hs}$) & 0.692/0.697 & 21.87/21.97 \\
    \hline
    \, \textbf{GPT-2$_{KN}$} &&\\
    \, \, w/ $Q_{hs}$ & 0.723/0.719 & 8.13/7.97 \\ 
    \, \, w/ $Q_{gen}$ & 0.728/0.720 & 7.48/6.34 \\
    \, \, w/ $Q_{hs \cup gen}$ & 0.735/0.740 & 6.30/5.81 \\
    \, \, w/ $Q_{hs \cup cn}$ & 0.727/0.731 & 7.17/6.16 \\
    \hline    
    \, \textbf{GPT-2$_{KN,MT}$} &&\\
    \, \, w/ $Q_{hs}$ & 0.744/0.748 & 11.69/10.24 \\ 
    \, \, w/ $Q_{gen}$ & 0.731/0.750 & 10.37/9.25 \\
    \, \, w/ $Q_{hs \cup gen}$ & 0.747/0.747 & 7.59/8.41 \\
    \, \, w/ $Q_{hs \cup cn}$ & 0.731/0.728 & 9.56/11.08 \\
    \hline
    \, \textbf{XNLG} && \\
    \, \, w/ $Q_{hs}$ & \textbf{0.824}/\textbf{0.828} & 14.42/6.22 \\ 
    \, \, w/ $Q_{gen}$ &  0.819/0.821 & 6.88/\textbf{4.03} \\
    \, \, w/ $Q_{hs \cup gen}$ &  0.812/0.817 & 6.98/5.19\\
    \, \, w/ $Q_{hs \cup cn}$ &  0.819/0.816 & \textbf{5.69}/4.64 \\
    \hline
  \end{tabular}
  \caption{Results of unique \texttt{HS} test set (U) for novelty and diversity. For comparison, we keep the results of whole test set (W).
  }\label{tab:unique_test}
\end{table}

\subsection{Human Evaluation Instruction} \label{sec:evaluation_instruction}
Three experts are tasked with \texttt{CN} generation evaluation. We briefed the annotators the aim of the evaluation, and then we explained common and appropriate \texttt{CN}s along with 10 examples of \texttt{HS}-\texttt{CN} pairs. The annotators are instructed to evaluate the generation with intuition. The generation is evaluated in terms of 3 aspects: (i) \textbf{suitableness:} How suitable the counter narrative is under evaluation for its hate speech, i.e. coherence and relevance of a counter narrative with regard to the hate speech. (ii) \textbf{informativeness:} How informative/specific the counter narrative is as a response for its hate speech. (iii) \textbf{intra-coherence:} How coherent the counter narrative is under evaluation as a response, i.e. internal coherence of a counter narrative regardless of its hate speech. In Table~\ref{tab:Annotation_examples} we give the annotation examples with the score explanations provided to the annotators.

\begin{table}[ht!]
\small
  \centering
  \begin{tabular}{p{0.95\linewidth}}
    \hline
    \textbf{HS:} Muslims do not want to assimilate. They demand that everyone shows respect for Islam, but they don't respect our laws. One day they will oppress us with the Sharia and polygamy. \\
    \hline 
    \textbf{CN:}  Do you really believe that they are a problem? I have a job.\\
    \textbf{evaluation}: suit.: 3, info.: 1, cohe.: 1 \\
    \textbf{Note}: The first sentence is relevant but not the second one. There is no coherence between 2 sentences. \\
    \hline
    \textbf{CN:}  As far as they do not force us to follow sharia laws, we should not be concerned with them following their laws. \\
    \textbf{evaluation}: suit.: 5, info.: 3, cohe.: 5 \\
    \textbf{Note}: The \texttt{CN} is an appropriate and coherent response, but not providing a substantial amount of information or facts. \\
    \hline
    \textbf{CN:}  Muslims should not be forced to assimilate, since it is not right and no one wants that. And polygamy is illegal and forbidden in UK and Muslims actually respect this ban. \\
    \textbf{evaluation}: suit.: 5, info.: 5, cohe.: 5 \\
    \textbf{Note}: The whole \texttt{CN} is perfectly suitable for the given \texttt{HS}. It is highly informative with the facts and internally coherent. \\
    \hline
  \end{tabular}
  \caption{Annotation examples with score explanations. }\label{tab:Annotation_examples}
\end{table}

\clearpage
\subsection{Generation Examples} \label{sec:sample_generation}

\begin{table}[ht!]
  \centering
  \begin{tabular}{p{15.5cm}}
    \hline
\textbf{HS:} Jews are Christ Killers. \\
\hline
\textbf{Human:} While the Catholic Church has for decades taught that Jews weren't collectively responsible, the argument laid out by the Pope Benedict XVI was a landmark statement. In his book, he not only confirms church teaching refuting the decide charge, he also explains biblically and theologically why there is \textcolor{blue}{no} basis in Scripture for the argument that \textcolor{blue}{the Jewish people} as a whole \textcolor{blue}{were responsible for Jesus' death.} \\
\hline
\hline
\textbf{TRF:} Why do you consider muslims like beasts? Did they do something wrong? If you take into consideration the crimes committed by muslims, why do you distinguish the muslim criminals from the non muslim? They are all criminals and all of them are punished. \\
\hline
\textbf{GPT-2:} I know many Muslims who are peaceful and respectful and do not kill. \\
\hline
\textbf{Candela:} Do you have proof? Where did you get this information from? I do not see the link between islamism and the news of the muslim community. \\
\hline
\textbf{GPT-2$_{KN}$:} No, the crucifixion of Jesus Christ is not a source for \textcolor{blue}{anti-Semitism}. \textcolor{blue}{The Jewish people were not responsible for the death of Jesus.} \\
\hline
\textbf{GPT-2$_{KN,MT}$:} On the basis of christianity, as christianity is not a religion, \textcolor{blue}{jews} are \textcolor{blue}{not responsible for} their own \textcolor{blue}{death}. \\ 
\hline
\textbf{XNLG:} \textcolor{blue}{In "jesus of nazareth-part ii" excerpts} the church's teaching that refuted the deicide charge of the \textcolor{blue}{jewish people}, as well as theologically explains one of the most controversial issues in christianity.\\ 
\hline
\textbf{KN:} Pope Benedict XVI has made a sweeping exoneration of the Jewish people \textcolor{blue}{for the death of Jesus} Christ. 
While the Catholic Church has for five decades taught that Jews weren't collectively responsible, Jewish scholars said... 
would help fight \textcolor{blue}{anti-Semitism} today. 
\textcolor{blue}{In "Jesus of Nazareth-Part II" excerpts} released Wednesday, Benedict explains biblically and theologically why there is \textcolor{blue}{no} basis in Scripture for the argument that \textcolor{blue}{the Jewish people} as a whole \textcolor{blue}{were responsible for Jesus' death.} \\
    \hline 
  \end{tabular}
  \caption{\parbox{15cm}{Sample cross-domain \texttt{CN} generation from different models.}}
 \label{tab:CN-app1}
\end{table}

\clearpage

\begin{table}[htbp]
  \begin{tabular}{p{15.5cm}}
    \hline
\textbf{HS:} Why gays do not try to heal? \\
\hline
\textbf{Human:} Contrary to some insinuations, sexual orientation \textcolor{blue}{is not} like \textcolor{blue}{a disease}. \\
\hline
\hline
\textbf{TRF:} Type imam boussena and we ll see who the scum is. \\
\hline
\textbf{GPT-2:} So they're the ones who are trying to make us hate them. \\
\hline
\textbf{Candela:} You see more and more attacks on muslims as a result of the controversies? Do you really want to believe in a `caliphate'? \\
\hline
\textbf{GPT-2$_{KN}$:} We know that homosexuality \textcolor{blue}{is not a disease}. \\
\hline
\textbf{GPT-2$_{KN,MT}$:} The genetics of homosexuality have been studied, but it seems to me that the human race is not suited to do that.\\
\hline
\textbf{XNLG:} And , \textcolor{blue}{contrary to} the \textcolor{blue}{insinuation} , \textcolor{blue}{sexual orientation is not like a disease . more recent studies have identified other potential links between homosexuality and genetics , and scientists have offered explanations on how same sex attraction can advance procreation , suggesting} that the same twin \textcolor{blue}{is more likely to be gay} , but finding the biological origins of the same \\
\hline
\textbf{KN:} Contrary to Perry's \textcolor{blue}{insinuation, sexual orientation is not like a disease. More recent studies have identified other potential links between homosexuality and genetics, and scientists have offered explanations on how same-sex attraction could advance procreation}. Studies of identical twins, who share the same genes, have shown that, if one twin is gay, the other \textcolor{blue}{is more likely to be gay, suggesting} sexual orientation is influenced by genetics. Scientific studies have demonstrated that biology plays a role in one's sexual orientation. \\

    \hline 
  \end{tabular}
  \caption{\parbox{15cm}{Samples of cross-domain \texttt{CN} generation from different models.}}
 \label{tab:CN-app2}
\end{table}

\clearpage

\begin{table*}[ht!]
  \centering
  \begin{tabular}{p{15.5cm}}
    \hline
\textbf{HS:} It seems that I cannot define Islam as a cult which promote wars and conquers host nations and legitimate this by its scriptures. Can I define them as narrow-minded, prejudiced, sexist and male chauvinist? \\
\hline
\textbf{Human:} Everyone has the right of free speech, while it does not mean anyone can use it to hurt people. Plus, based on what you critisize Islam? I recommend you to read something about true Islam and Muslims. You would find that they are no difference from us. \\
\hline
\hline
\textbf{TRF:} Islam is all about the religions and tolerance insert quote from koran. Migrants is not evil? \\
\hline
\textbf{GPT-2:} It seems to me that this is the antithesis of what Islam teaches. \\
\hline
\textbf{KN $Q_{hs}$:} \textcolor{blue}{Hakim al-Mutairi}, secretary-general of the emirate's Umma ( Islamic \textcolor{blue}{Nation} ) \textcolor{blue}{party}, \textcolor{blue}{urged the pope to} apologise \textcolor{blue}{immediately "to the Muslim world for his calumnies against the Prophet Muhammad and Islam}".... 
Instead of bringing Islam and Christianity closer, he is straining relations between the two religions". 
\textcolor{blue}{In the present political atmosphere such views can be exploited by those} who are trying to malign Muslims and Islam. \\
\textbf{Candela:} I have no accountability. I have no problem with those who claim to belong to, but i do not have the right to practice whatever religion. \\
\textbf{GPT-2$_{KN}$:} The Qur'an clearly teaches peace, love and tolerance towards everyone, including the non-Muslims. The only thing that Islam does not have is a message of conquest and war.\\
\textbf{GPT-2$_{KN,MT}$:} Islam is a religion founded on the principle that you must believe in the good will of all people. \\
\textbf{XNLG:} Extremist muslim national party secretary, \textcolor{blue}{hamid al mutairi, urged the pope to} define \textcolor{blue}{immediately to the muslim world for his calumies against the prophet muhammad and islam}. Pope benedict xvi was full of hostility and hatred against muslims and islam. \textcolor{blue}{In the present political atmosphere such views can be exploited by those who} bring \\
\hline
\textbf{KN $Q_{hs \cup gen}$:} \textcolor{blue}{Islam, as practiced by the vast majority of} people, is \textcolor{blue}{a peaceful religion}, \textcolor{blue}{a religion that respects others}. "The large \textcolor{blue}{majority of Muslims} tinker," said Franck Fregosi, a sociologist who has written extensively on Islam in Europe....\\
\textbf{GPT-2$_{KN}$:} Hi, I do not understand the problem you have. \textcolor{blue}{The vast majority of Muslims} do not advocate wars and conquer the world. They practice \textcolor{blue}{a peaceful religion that respects others}.\\
\textbf{GPT-2$_{KN,MT}$:} It seems that i have just come across the iota of \textcolor{blue}{muslims} who have been insulted, persecuted, killed, raped, harassed, murdered, killed by the muslims in europe. \\
\textbf{XNLG:} It seems that \textcolor{blue}{islam}, \textcolor{blue}{as practiced by the vast majority of people, is a peaceful religion, a religion that respects others. The large majority of muslims tinker}, as \textcolor{blue}{franck fregosi, who has written extensively on islam in europe}. It seems that it does not define islam as an inclusion and respect for all of the religion \\
\hline
\textbf{KN $Q_{hs \cup cn}$:} No doubt, \textcolor{blue}{it is the legitimate right of Muslims to build a community center near} Ground Zero. So please stop conflating \textcolor{blue}{criticism of Islam with discrimination of Muslims. As a nation of free people, we can never defeat Islam but we can make war against terrorists and governments that support them}. a "moderate Islamic countries" when all unanimously reject the Universal Declaration of Human Rights and support The Cairo Declaration of Human Rights in Islam. \\
\textbf{GPT-2$_{KN}$:} The Islamic countries do not promote wars and conquer host nations. Islam is not a cult but a peaceful religion which respects all peoples.\\
\textbf{GPT-2$_{KN,MT}$:} The vast majority of men in the world are not able to articulate the spiritual values of islam.\\
\textbf{XNLG:} It seems that \textcolor{blue}{it is the legitimate right of muslims to build a community center near} zero ground, so please confine \textcolor{blue}{criticism of islam with discrimination against muslims. As a nation of free people, we can never defeat islam, but we can make war against terrorists and governments that support them}. My comment gives a lot of information about what islam and muslims \\
    \hline 
  \end{tabular}
  \caption{\parbox{15cm}{Sample of in-domain generation.}}
 \label{tab:CN-app5}
\end{table*}

\end{document}